\documentclass[letterpaper, 10 pt, conference]{ieeeconf}  

\IEEEoverridecommandlockouts    
\usepackage{hyperref}
\usepackage{hhline}
\usepackage{soul}
\usepackage{xcolor}
\usepackage{cite}
\usepackage{import}
\usepackage{amsmath}
\usepackage{graphicx}
\usepackage{amsfonts} 
\usepackage{bm}
\usepackage[capitalise]{cleveref}
\usepackage{booktabs}
\usepackage{authblk}

\newcommand{\etal}{\textit{et al}. }
\newcommand{\ie}{\textit{i}.\textit{e}., }
\newcommand{\eg}{\textit{e}.\textit{g}. }

\title{\LARGE \bf
Dynamic object goal pushing with mobile manipulators through model-free constrained reinforcement learning
}

\author{Ioannis Dadiotis$^{1,2}$, Mayank Mittal$^{3,4}$, Nikos Tsagarakis$^{1}$, Marco Hutter$^{3}$

\thanks{\hspace{-12pt} This project has received funding from the European Union’s Horizon Europe Framework Programme under grant agreement No 101070596 (euROBIN), the SNSF NCCR dfab, and an ETH RobotX research grant funded through the ETH
Zurich Foundation. Contact: {\tt\small ioannis.dadiotis@iit.it}}
\thanks{$^{1}$HHCM lab, IIT, Genoa 16163, Italy}
\thanks{$^{2}$DIBRIS, University of Genoa, Genoa 16145, Italy.}
\thanks{$^{3}$RSL, ETH Z\"{u}rich, Z\"{u}rich 8092, Switzerland. $^{4}$NVIDIA.}
}

\usepackage{tikz}
\usepackage{textcomp}
\usepackage{lipsum}
\usepackage{outlines}

\usepackage{multirow}   

\newcommand\copyrighttext{%
  \scriptsize \textcopyright 2025 IEEE. Personal use of this material is permitted.
  Permission from IEEE must be obtained for all other uses, in any current or future
  media, including reprinting/republishing this material for advertising or promotional
  purposes, creating new collective works, for resale or redistribution to servers or
  lists, or reuse of any copyrighted component of this work in other works.}
\newcommand\copyrightnotice{%
\begin{tikzpicture}[remember picture,overlay]
\node[anchor=south,yshift=10pt] at (current page.south) {\parbox{\dimexpr\textwidth-\fboxsep-\fboxrule\relax}{\copyrighttext}};
\end{tikzpicture}%
}

\usepackage{array}
\usepackage{color, colortbl}

\newcolumntype{P}[1]{>{\centering\arraybackslash}p{#1}}

\begin{document}
\bstctlcite{IEEEexample:BSTcontrol}

\maketitle
\copyrightnotice

\thispagestyle{empty}
\pagestyle{empty}

\begin{abstract}
Non-prehensile pushing to move and reorient objects to a goal is a versatile loco-manipulation skill. In the real world, the object's physical properties and friction with the floor contain significant uncertainties, which makes the task challenging for a mobile manipulator. In this paper, we develop a learning-based controller for a mobile manipulator to move an unknown object to a desired position and yaw orientation through a sequence of pushing actions. The proposed controller for the robotic arm and the mobile base motion is trained using a constrained Reinforcement Learning (RL) formulation. We demonstrate its capability in experiments with a quadrupedal robot equipped with an arm. The learned policy achieves a success rate of 91.35\% in simulation and at least 80\% on hardware in challenging scenarios. Through our extensive hardware experiments, we show that the approach demonstrates high robustness against unknown objects of different masses, materials, sizes, and shapes. It reactively discovers the pushing location and direction, thus achieving contact-rich behavior while observing only the pose of the object. Additionally, we demonstrate the adaptive behavior of the learned policy towards preventing the object from toppling.
\end{abstract}

\section{Introduction}
Moving and reorienting heavy or bulky objects along large and complex real-world pathways requires combining mobility and manipulation. This task is achievable through non-prehensile pushing actions without requiring a dedicated gripper or the need to grasp a handle on the object. In real-world scenarios, however, the object and terrain's physical properties (\eg mass, size, friction coefficient) are typically unknown and can reduce a controller's performance. Additionally, non-prehensile pushing interaction may yield relative motion between the robot and object at the contact point, \eg contact sliding or relative rotation. This motion necessitates a controller capable of reactively adapting the pushing location and direction by dynamically breaking and making contact at new locations with the object. We refer to this contact-rich behavior as contact switching.

Achieving online contact switching behavior during pushing is challenging with model-based techniques \cite{hogan_hybrid_mpc, moura_to_planar_manmipulation} or controllers that rely solely on force/tactile feedback \cite{force_push_heins, ozdamar2024pushing}. As a result, recent works leverage Reinforcement Learning (RL) to address contact switching~\cite{ferrandis2023nonprehensile} and demonstrate notable robustness against unknown objects~\cite{quadruped_pushing}. Despite these achievements, the method in \cite{ferrandis2023nonprehensile} is limited to fixed-base manipulators pushing a lightweight, small object on a table. Jeon~\etal\cite{quadruped_pushing} achieve object pushing with the base of a mobile robot, without an arm. In both cases, the policies generate only 2D planar motion commands, which are insufficient for manipulating objects that are prone to toppling (\eg objects with a thin base, large CoM height, or high friction coefficient flooring). To address this limitation, we focus on interacting with objects by pushing them to different 3D locations on their surface, enabling more versatile and stable manipulation.
\begin{figure}
  \centering
  \graphicspath{{figures/}}
  \includegraphics[width=0.8\columnwidth]{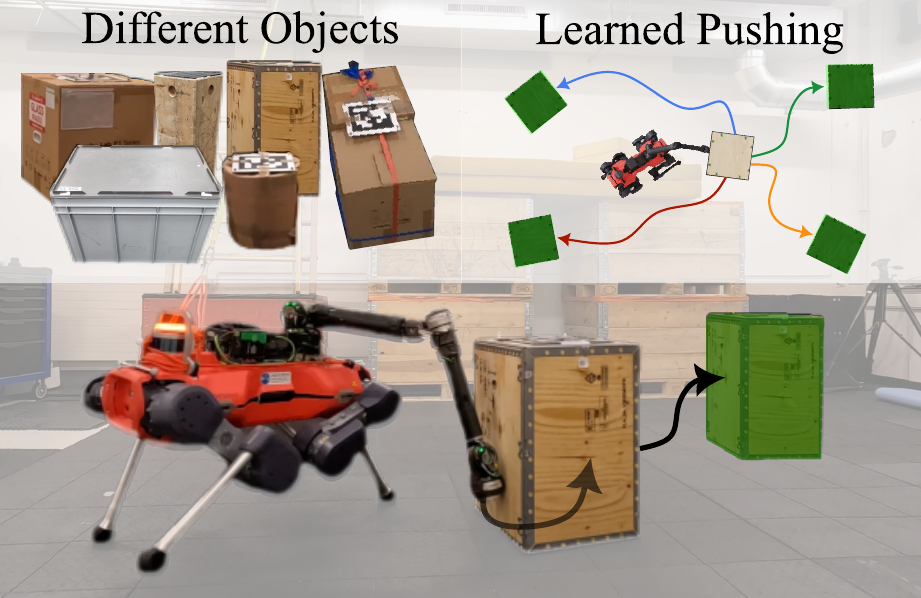}
  \vspace{-7pt}
  \caption{Dynamic object pushing with a quadrupedal manipulator. The proposed controller learns to push unknown objects towards different goals. The motions are included in the supplementary video \href{https://youtu.be/wGAdPGVf9Ws?si=j9YNlEufzQIGlPz4}{(link)}.}
  \label{Fig:fig1}
\end{figure}

In this work, we present a learning-based controller for a mobile manipulator to dynamically move and reorient unknown objects using non-prehensile pushing actions. We tackle the task complexity by using a state-of-the-art constrained RL algorithm \cite{chanesane2024cat} that minimizes reward engineering efforts and can satisfy the various constraints of the task, \eg arm actuator limits, self-collisions. The policy's action space consists of cartesian commands for the base and joint-space commands for the mounted articulated arm; thus, we directly control the arm in joint space. Our proposed approach achieves robustness against unknown objects and learns online contact switching to push the object to various planar goal poses. The resulting behavior demonstrates the adaptability of the pushing location, which is crucial for avoiding object toppling. Our key contributions are:
\begin{itemize}
    \item We develop a learning-based controller for mobile manipulators to repose objects on a plane through pushing. Importantly, our approach incorporates object balance as a key task requirement, which has not been addressed in prior works.
    \item Our controller demonstrates robustness to unknown objects with differing physical properties, such as mass, material, size, and shape. It autonomously handles contact switching by dynamically identifying the contact points and adjusting the push direction. Even though the policy only observes the object's pose, it adapts to its xy-footprint and lowers the pushing location for thinner objects to maintain stability. 
    \item We validate our approach in simulation and real-world experiments with a quadrupedal manipulator, achieving consistent success across diverse scenarios, including high-friction surfaces and thin, easily toppled objects.
\end{itemize}

\section{Related work}
\label{sec:related_work}
\subsection{Dynamic mobile manipulation control}
\label{sec:mobile_manipulation}
%
Model-based approaches, such as Trajectory Optimization (TO) \cite{dadiotis2022trajectory, polverini_pushing} and Model Predictive Control (MPC) \cite{sleiman_unified_mpc, minniti_mpc, mayank_articulated, dadiotis_wholebody, pankert_mpc}, are frequently used to control dynamic mobile manipulators. A major limitation is that their real-time performance requires a predefined dynamically feasible contact schedule\cite{sleiman_science_robotics}, which is computationally expensive and is typically done offline. Thus, these approaches are insufficient to achieve online contact switching. Another drawback is that they rely on a model of the robot and its environment.

On the other hand, model-free RL-based controllers\footnote{We use the term "model-free" when a model and its derivatives are not required in the controller structure. A model may still be used for simulation.} can achieve robustness against uncertainties via domain randomization in simulation-based training. Several works have successfully used this approach to control the end-effector of mobile manipulators \cite{fu2023deep, liu2024visual, ha2024umilegs}. However, they mainly tackle the tracking problem in free space and use the resulting controller to grasp lightweight objects. Thus, it is not clear how these approaches can scale to contact-rich tasks with heavy object interaction.
\subsection{Non-prehensile pushing motion control}
\label{sec:nonprehensile_manipulation}

Various model-based \cite{hogan_hybrid_mpc, moura_to_planar_manmipulation, inaba_humanoid_pushing, polverini_pushing} and learning-based approaches \cite{ferrandis2023nonprehensile, peng_rl, Lin_vision-proprioception, rizzardo_push_cube} have been tailored to robot pushing behaviors. However, these methods tend to suffer from the inherent limitations mentioned in \cref{sec:mobile_manipulation}. Ferrandis~\etal~\cite{ferrandis2023nonprehensile} achieve contact switching using RL with a categorical action distribution. Despite this achievement, this approach has been limited to a fixed-base manipulator, objects with negligible mass and small size on a low-friction surface.

The works of \cite{force_push_heins, ozdamar2024pushing} propose a force and tactile feedback-based approach, respectively, for statically pushing with a mobile base robot without achieving contact switching (since the feedback signal is lost at a contact break). Moreover, they only consider moving an object to a goal position and not reorienting it. Jeon~\etal\cite{quadruped_pushing} train an RL policy for guiding a quadrupedal locomotion controller towards pushing diverse objects with the robot base. Their controller generates planar 2D actions for the base motion, and they do not consider object toppling as a possibility in their scenarios. In contrast, we evaluate our approach including cases where the object can topple (\ie object with a small xy-footprint on high-friction flooring). By generating 6D commands for the mobile base and joint commands for the arm, our controller can push to different locations on the object's surface.

\section{Method}
\label{sec:method}
\begin{figure}
  \centering
  \graphicspath{{figures/}}
  \includegraphics[width=\columnwidth]{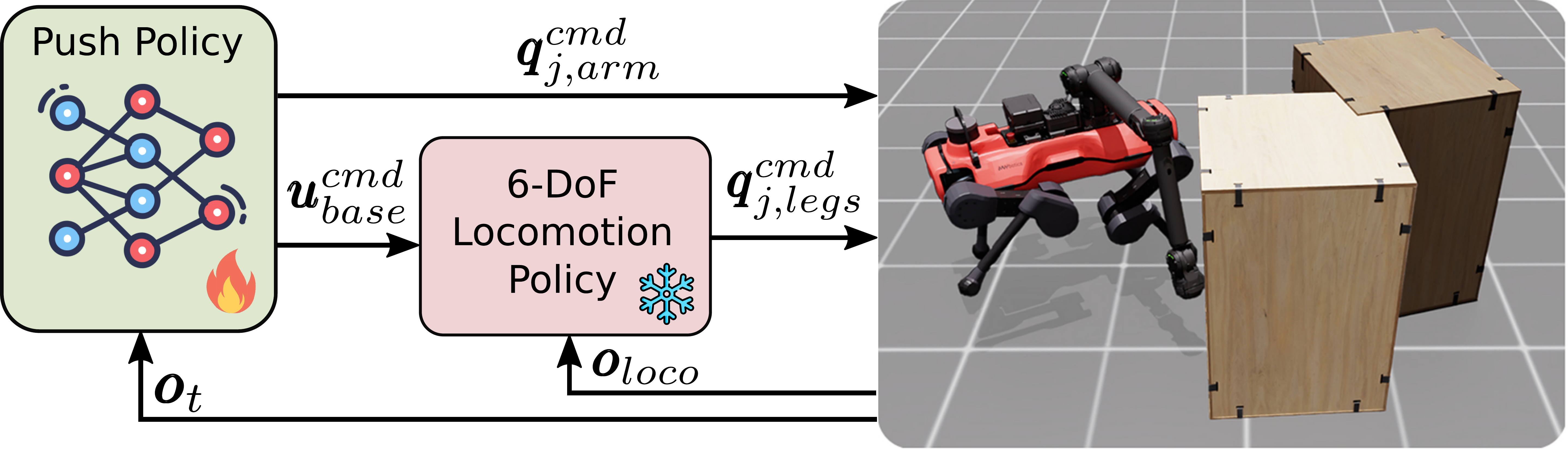}
  \vspace{-20pt}
  \caption{The control pipeline used for moving and reorienting an object to a planar goal (darker object). Push policy is the proposed controller.}
  \label{Fig:framework}
\end{figure}

We train a push policy for mobile manipulators to repose objects on a plane. The policy provides cartesian commands for the mobile base and joint position commands for the first five joints of the arm. We freeze the 6\textsuperscript{th} joint since it is only useful when using a gripper. The velocity commands are sent to a pre-trained locomotion controller to convert them into joint position commands for the legs. Both the push policy and locomotion policy are inferred at the same frequency of 50 Hz. The overall architecture is shown in~\cref{Fig:framework}.
\subsection{Locomotion control}
\label{subsec:robot_locomotion}
The approach is validated using a quadrupedal mobile platform, ANYmal, with a six DoF robotic arm mounted on it. The locomotion controller is a student policy similar to the one in \cite{miki2024learningwalkconfinedspaces}. It accepts the base command $\pmb u_{base}^{cmd} = (v_x, \ v_y, \ \omega_z, \ \zeta, \ \theta_, \ h) \in \mathbb{R}^6$ and outputs leg joint position targets. The six components of the command $\pmb u_{base}^{cmd}$ consist of linear velocity in $x$ and $y$ directions, angular yaw velocity, roll and pitch angle, and height position, respectively. The policy is trained with randomized arm motions and includes the arm joint positions in the observation. This way, the resulting locomotion policy is robust against the range of arm motions. While training the proposed controller (push policy), we freeze the pre-trained locomotion controller.

\subsection{RL goal pushing environment}
\label{sec:rl_environment}

We implement the task of moving and reorienting an object using NVIDIA Isaac Lab \cite{mittal2023orbit} for training the RL policy with 4096 parallel simulated robots for 20000 iterations. We modified the Proximal Policy Optimization (PPO) implementation from \cite{rudin2022learning} with the changes in \cite{chanesane2024cat} to derive the constrained PPO formulation. The reader is advised to read \cite{chanesane2024cat} for the details on constrained RL.

\textit{Notation}: In the following we use $\pmb p \in \mathbb{R}^3$, $\pmb R \in \mathbb{SO}(3)$, $\pmb v \in \mathbb{R}^3$ and $\pmb \omega \in \mathbb{R}^3$ to denote position, rotation matrix, linear and angular velocity of a body's frame, respectively. The body frame name is denoted as a right subscript and the reference frame as a left superscript (omitted when the reference frame and body frame are the same). We use the letters $w$, $b$, $o$, $g$, and $e$ to refer to the world, robot base, object, object goal, and arm end-effector frames, respectively. For the relative position between two body frames, two letters are used at the right subscript, \eg $^b\pmb p_{oe}$ is the relative position of the end-effector w.r.t. the object frame expressed in the robot base frame. We use $\widehat{\cdot}$ and $\|\cdot\|$ to express a given vector's unit vector and length, respectively.

\begin{figure}
  \centering
  \graphicspath{{figures/}}
  \includegraphics[width=0.9\columnwidth]{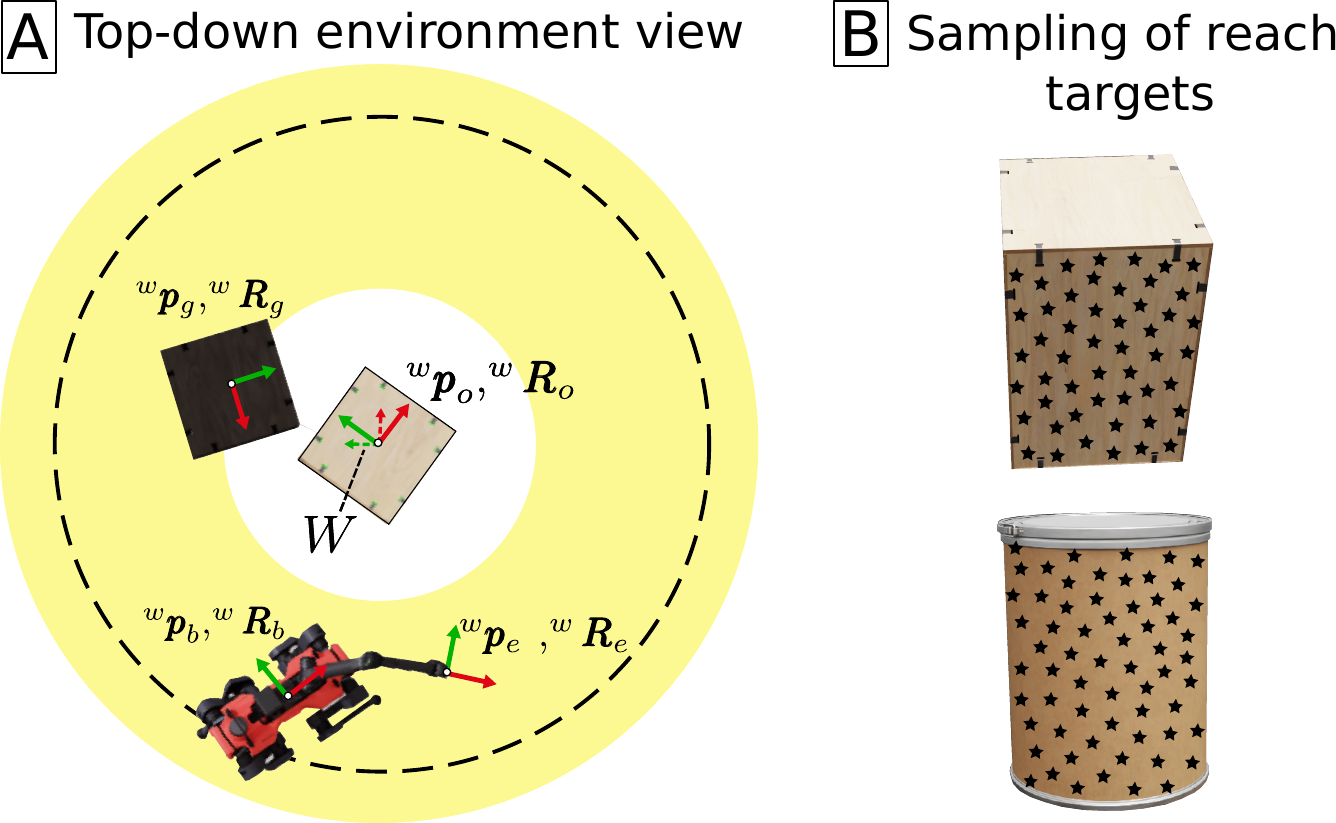}
  \vspace{-8pt}
  \caption{A) The object's position is set to the environment origin ($W$), the robot base position is randomly sampled within an origin-centered annulus (yellow-shaded area), and the object goal (dark rectangle) within a circular area (dashed line). The robot, object, and goal are spawned with a random yaw orientation. B) Sampling on the object surface encourages interaction with different parts of the object during training.}
  \label{Fig:environment}
\end{figure}

\textit{Environment \& commands}: The RL training environment consists of multiple object-centered environments (parallelly simulated), each reset when there is an episode timeout (20 sec after the last reset) or when there is an unrecoverable object or robot fall. \cref{Fig:environment}A shows a single environment after a reset when the robot, object, and object goal positions are resampled in polar coordinates. The object ($^w\pmb p_o, ^w\pmb R_o$) is spawned at the origin of the environment ($W$). In contrast, the robot is spawned with its base frame ($^w\pmb p_b, ^w\pmb R_b$) at a random position inside an origin-centered annulus (yellow area with radius in the range $[1.2, 2.5] \ m$). An annulus is selected to prevent the robot from starting too close to the object or having collisions between them. Moreover, we aim for the robot to learn an approaching motion. The object's goal position $^w \pmb p_g$ is sampled at an origin-centered circular area (black dashed line, radius $2 \ m$). The yaw orientation of the object, the object goal, and the robot base are sampled randomly along the whole range $[-\pi, \pi]$. We consider success when the distance between the object's frame and the goal is less or equal to $d_{success} = 10 \text{ cm}$ and the angle between their orientation is less or equal to $\theta_{success} = 10 \text{ deg}$.

During exploration in simulation, we want to encourage interactions of the robot with the whole surface of the object so that the RL agent discovers which part of the object is better to interact with. To that end, a reaching target position $^w\pmb p_r$ is randomly sampled on the object's vertical surfaces, as shown in \cref{Fig:environment}B, at each environment reset. This target is used in the reward function to guide the robot EE towards interacting with the object, as explained in \cref{sec:rewards_constraints}.

\subsection{Observation \& action space}
\label{sec:obsrevation_action}
This work uses an asymmetric actor-critic approach \cite{pinto2017asymmetric, ma2023learning} where the critic can access privileged information available only in simulation and noiseless. All observed quantities are described in \cref{tab:observations}. It is worth noting that the only information regarding the object included in the actor's observation vector $\pmb o_t \in \mathbb{R}^{54}$ is the object pose and, thus, the deployed policy has no knowledge about the object size and dynamics. The action vector $\pmb a_t = (\Delta \pmb u_{base}^{cmd}, \ \Delta \pmb q_j^{cmd}) \in \mathbb{R}^{11}$ consists of base commands $\Delta \pmb u_{base}^{cmd}$ for the locomotion policy described in \cref{subsec:robot_locomotion} and arm joint position commands $\Delta \pmb q_j^{cmd} \in \mathbb{R}^5$. The actions generated by the policy refer to deviations from a default base state (zero velocities, zero orientation, and default height of 0.5 m) and a default arm configuration, respectively. Thus, they are transformed into absolute values before being passed on to the locomotion policy and low-level joint impedance controllers. In \cref{tab:observations}, the quantities observed by the critic $\pmb o_t^{critic} = (\pmb o_t, \ \pmb o_t^{pr}) \in \mathbb{R}^{73}$, including the privileged information $ \pmb o_t^{pr}$.

\begin{table}
\caption{Observations for the actor ($\pmb o_t$) and critic ($\pmb o_t, \pmb o_t^{pr}$). Unlike the actor, the critic receives noiseless observations.}
\label{tab:observations}
\vspace{-18pt}
\begin{center}
  \begin{tabular}{P{0.25cm}|P{0.55cm}|p{4.25cm}|P{0.35cm}|P{1.1cm}}
  \hline
   \multicolumn{3}{P{5.1cm}|}{\textbf{Description}}                                                          & \textbf{Dim}      & \textbf{Noise}            \\
   \hhline{=|=|=|=|=}
   \multirow{10}{*}{$\pmb o_t$} & $^b\pmb p_{oe}$            & EE-object relative position w.r.t. base       & 3        & $\mathcal{U}(\pm0.02)$         \\ \cline{2-5}
   & $^b\pmb R_o$                & object rotation matrix w.r.t. base            & 9        & $\mathcal{U}(\pm0.01)$         \\ \cline{2-5}
   & $\Delta \pmb q_j$           & arm joint position readings w.r.t. default configuration               & 5        & $\mathcal{U}(\pm0.01)$         \\ \cline{2-5}
   
   & $\pmb v_b$                & robot base linear velocity                 & 3        & $\mathcal{U}(\pm0.01)$         \\ \cline{2-5}
   & $\pmb \omega_b$           & robot base angular velocity                & 3        & $\mathcal{U}(\pm0.20)$          \\ \cline{2-5}
   & $\dot{\pmb q}_j$            & arm joint velocity readings                & 5        & $\mathcal{U}(\pm0.50)$          \\ \cline{2-5}
   & $^b\pmb g_z$                & projected gravity unit vector & 3        & $\mathcal{U}(\pm0.05)$         \\ \cline{2-5}
   & $^b\pmb p_{og}$             & object-goal relative position w.r.t. base   & 3        & $\mathcal{U}(\pm0.02)$         \\ \cline{2-5}
   & $^o\pmb R_g$                & goal orientation w.r.t. object              & 9        & $\mathcal{U}(\pm0.01)$         \\ \cline{2-5}

   & $\pmb a_{t-1}$              & previous actions                           & 11       & -                    \\
   \hhline{=|=|=|=|=}
   \multirow{8}{*}{$\pmb o_t^{pr}$} & $\lambda_{e}$              & EE-object contact state                    & 1       &                       \\ \cline{2-4}
   & $^b\pmb p_{com}$            & object CoM position w.r.t. robot base         & 3       &                       \\ \cline{2-4}
   & $m$                         & object mass                                & 1       &                       \\ \cline{2-4}
   & $\pmb d$                    & object dimensions                   & 3       & -                     \\ \cline{2-4}
   & $\pmb I_o$                  & object's principal moments of inertia      & 3       &                       \\ \cline{2-4}
   & $^b\pmb v_{o}$            & object linear velocity w.r.t. robot      & 3       &                       \\ \cline{2-4}
   & $^b\pmb \omega_{o}$       & object angular velocity w.r.t. robot     & 3       &                       \\ \cline{2-4}
   & $\pmb \kappa_{sh}$          & one-hot vector for object shape            & 2       &                       \\ \hline    
   \end{tabular}
\end{center}
\end{table}

\begingroup
\renewcommand{\arraystretch}{1.3}
\begin{table*}
\caption{Rewards and constraints used for training along with the initial and final values of the constraint hyperparameter $p_{i}^{\lowercase{max}}$ from CaT~\cite{chanesane2024cat} and their respective curriculum schedule over the learning iterations.}
\label{tab:constraints}
\vspace{-12pt}
\begin{center}
  \begin{tabular}{p{4cm}||p{2.7cm}|p{5.5cm}|P{0.4cm}|P{1.4cm}|P{1.0cm}}
  \hline
    \multicolumn{1}{P{4cm}||}{\textbf{Rewards}} & \multicolumn{5}{P{12.0cm}}{\textbf{Constraints}} \\ 
   \hhline{=||=|=|=|=|=}
   \multirow{2}{3.8cm}{$r_{1,t} = \exp\Big({-\frac{\|^w\pmb K_g - ^w\pmb K_o\|}{\sigma_3^2}}\Big)$} & \multicolumn{1}{P{2.7cm}|}{\textbf{Description}}                          & \multicolumn{1}{P{5.5cm}|}{\textbf{Formulation}} & \textbf{Dim} & $\pmb{p_{i}^\text{max}}$ & \textbf{Iterations} \\
   \cline{2-6}
    &base command limits                   & $\pmb c_{a}^{base} = \text{max}(\pmb u_{base}^{cmd} - \pmb u_{base}^{upper}, \pmb u_{base}^{low} - \pmb u_{base}^{cmd})$ & 6   & $0.01 \rightarrow 0.2$      &   \multirow{7}{1.0cm}{$0 \rightarrow 12 \cdot 10^3$} \\ \cline{1-5}
   \multirow{2}{3.8cm}{$r_{2,t} = \exp\Big({-\frac{\|^w\pmb p_{er}\|}{\sigma_1^2}}\Big)$} &arm command limits                    & $\pmb c_{a}^{arm} = \text{max}(\pmb q_j^{cmd} - \pmb q_j^{upper}, \pmb q_j^{lower} - \pmb q_j^{cmd})$      & 5   & $0.05  \rightarrow 0.9$ \\ \cline{2-5}
     & arm action rate limits               & $\pmb c_{\dot{a}}^{arm} = \frac{|\Delta\pmb q_{j, t}^{cmd} - \Delta\pmb q_{j, t-1}^{cmd}|}{dt} - \dot{\pmb q}^{lim} $     & 5   & $0 \rightarrow 0.05$    & \\ \cline{1-5}
   \multirow{3}{3.8cm}{$r_{3,t} = \exp\Big({\frac{^w\widehat{\pmb v}_o \cdot ^w\widehat{\pmb p}_{og}}{\sigma_2^2} - 1}\Big)$} &arm joint position limits            & $\pmb c_{q_j} = \text{max}(\pmb q_j - \pmb q_j^{upper}, \pmb q_j^{low} - \pmb q_j)$                               & 5   & $0.05  \rightarrow 0.9$     &   \\ \cline{2-5}
    &arm joint velocity limits            & $\pmb c_{\dot{q}_j} = |\dot{\pmb q}_{j}| - \dot{\pmb q}^{lim}$                                      & 5   & $0.05  \rightarrow 0.9$     &   \\ \cline{2-5}
   &arm joint torque limits              & $\pmb c_{\tau_j} = |\pmb \tau_j| - \pmb \tau_j^{lim}$                                         & 5   & $0 \rightarrow 0.015$   &  \\ \cline{1-5}
   \multirow{3}{3.8cm}{$r_{4,t} = \exp\Big({-\frac{|\Delta \pmb u_{base, t}^{cmd} - \Delta \pmb u_{base, t-1}^{cmd}|}{\sigma_{4,b}^2}}\Big) + \exp\Big({-\frac{|\Delta\pmb q_{j, t}^{cmd} - \Delta\pmb q_{j, t-1}^{cmd}|}{\sigma_{4,a}^2}}\Big) $} & leg joint torque limits              & $\pmb c_{\tau_{j,leg}} = |\pmb \tau_{j, leg}| - \pmb \tau_{j,leg}^{lim}$                                    & 12   & $0 \rightarrow 0.01$    &   \\ \cline{2-6}
    & undesired robot-object \& self-collisions    &  $c_{coll} = \begin{cases}
                                                                        1 & \text{, if a collision occurs}, \\
                                                                        0 & \text{, otherwise}.
                                                                     \end{cases}$                                                       & 18       & 1.0              &  \multirow{2}{1.0cm}{No curriculum}           \\ \cline{2-5}
   &object balance                      & $c_{\theta_{obj}} = \begin{cases} 
                                                    |\theta| - \theta^{lim} & \text{, if } \|^b\pmb v_b\| > 0, \\
                                                    0 & \text{, otherwise}.
                                                 \end{cases}$                                                                           & 1       & 0.25             &                                  \\
   \hline
   \end{tabular}
\end{center}
\end{table*}
\endgroup
\subsection{Rewards \& constraints}
\label{sec:rewards_constraints}
In this section, we describe the reward and constraint terms included in the training. We tune them manually to achieve convergence in simulation and then transfer the policy to the hardware zero-shot without further adjustments.

\textit{Rewards}: The total reward $r_t^{tot} = \sum_{n=1}^{4} w_i r_{i,t}$ consists of the sum of the reward terms shown in \cref{tab:constraints}. For the weights we used the values $w_1 = 2.5$, $w_2 = 1.25$, $w_3 = 0.156$, $w_4 = 0.3$. The term $r_{1,t}$ is the main task reward, which encourages minimizing the distance between the eight keypoints of the object and the keypoints of the goal, where these are defined as the vertices of the oriented bounding box of the object (similarly to \cite{allshire_trifinger, quadruped_pushing}). We denote the position of the keypoints of the object and its goal as $^w\pmb K_o \in \mathbb{R}^{24}$ and $^w\pmb K_g \in \mathbb{R}^{24}$, respectively. The reward term $r_{2,t}$ encourages the agent to minimize the distance between the arm EE and the reach target ($^w\pmb p_r$) sampled on the object's surface at each episode. The weight $w_2$ of this term is downscaled by a factor of 4 after 1500 learning iterations. We do this to encourage the robot EE to approach and interact with different parts of the object at the beginning, and we do not care about accurately reaching the sampled position. The term $r_{3,t}$ rewards object linear velocity with direction towards the object goal. We do not include the magnitude of the velocity in this term to avoid the robot pushing the object aggressively. Finally, the term $r_{4,t}$ comprises action rate regularization. If the task is successful, we increase the task reward to $r_{1,t}=2$, which is two times the maximum possible value of this term, so that the robot learns to achieve the specified tolerance instead of staying close to it. In case of success, the other rewards take values $r_{2,t}=r_{2,t-1}$, $r_{3,t}=0$ since once the goal has been reached, we do not encourage object velocity or interaction with the arm EE.
 
\textit{Constraints}: As proposed in \cite{chanesane2024cat}, we apply curriculum learning over most of the constraints by increasing the maximum probability for reward termination along training (linearly increasing $p_i^\text{max}$ from an initial smaller value to a final larger one after a number of learning iterations). In practice, at the beginning of the training, we encourage exploration by limiting the effect of constraint violation on the reward termination probability. We emphasize achieving strict constraint satisfaction for undesired collisions, arm joint position, and velocity limits. The undesired collisions include robot self-collisions and collisions of the object with the robot base, legs, and the arm's shoulder, upper arm, elbow, and forearm links. The arm joint torque constraint is not strict since we use as limits the maximum nominal torque values of the actuators and not the peak ones (which the joints can reach for short periods). We also include a constraint for the base command $\Delta \pmb u_{base}^{cmd}$ using as limits the ranges used during the training of the locomotion controller. Finally, the object balance constraint requires that the object's inclination angle be less than a specified threshold $\theta^{lim}=10^{\circ}$. All the constraints used for training, their dimension, and the curriculum applied are shown in \cref{tab:constraints}. 

\subsection{Domain randomization \& deployment}
\label{sec:domain_randomization}
To render the learned policy robust for deployment on the hardware, we randomize several factors in the simulated environment. The actor's observations are subject to additive uniform noise, as specified in \cref{tab:observations}. The static and dynamic friction between the object and the floor is randomized within the range of $[0.4,1.25]$ for the values of the combined coefficient. Moreover, the object mass is randomly sampled in the range 1-10 kg, and the center of mass (CoM) position is randomized around the object's centroid with deviation in the range $(dx, dy, dz)=([-0.25d_x, 0.25d_x], [-0.25d_y, 0.25d_y], [-0.6d_z, 0.25d_z])$, where $d_i$ is the dimension of the object in direction $i$. We also randomize the object dimensions in the ranges $(x, y, z)=([0.25, 0.75]m, [0.25, 0.75]m, [0.4, 1.0]m)$, and we train with cuboids and cylinders. Finally, the base mass for each simulated robot is randomly modified by $\pm$ 5 kg, and random pushes are applied to the robot base every 7 to 10 sec. The arm joint positions are randomized around the default configuration at each environment reset.

During deployment, we rely on an external motion capture system to get the object and robot base 6D pose information needed to derive the observation $\pmb o_t$. During testing on hardware, we infer the policy until success is achieved; then, we set the base command for the locomotion controller $\Delta \pmb u_{base}$ to zero to avoid the robot continuing stepping in place.

\begin{figure*}
  \centering
  \graphicspath{{figures/}}
  \includegraphics[width=\textwidth]{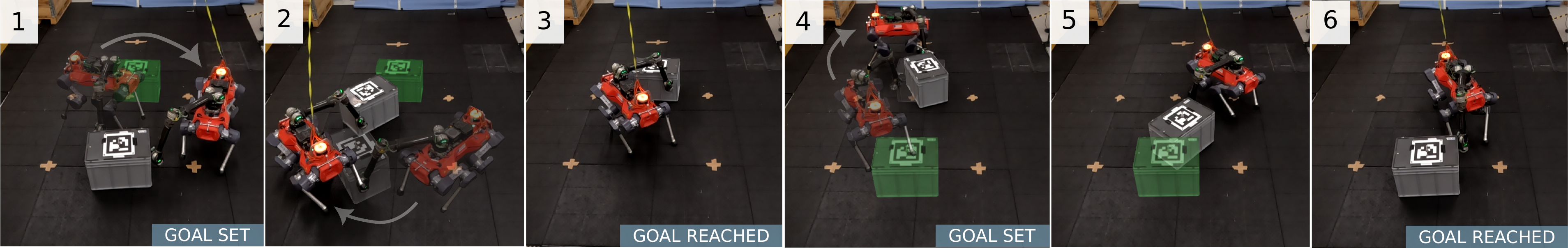}
  \vspace{-20pt}
  \caption{Experimental validation of the proposed controller for sequentially moving and re-orienting an object between two goal poses.  The robot pushes a plastic box of 6.4 kg from one goal to another. The goal poses are shown as green boxes. Snapshots of previous times are shown with lower opacity. The robot successfully goes around the object to push from the correct side towards the goal (1-2, 5).}
  \label{Fig:experiments_ptp}
\end{figure*}
\section{Results \& Evaluation}
\label{sec:results_main}
\subsection{Success rate \& object balance}
\label{sec:success_rate}
In this section, we present the achieved success rate of the proposed policy after simulating 4096 parallel environments for a single episode. We also provide details on the decision to include the object balance constraint $c_{\theta_{obj}}$ described in \cref{tab:constraints} and the sampling of reach targets $^w \pmb p_r$ on the object's surface. To that end, we conduct an ablation study in simulation by training the same policy without the balance constraint and by replacing the EE reach targets on the surface with the object's centroid. As shown in \cref{tab:success_rate_topple}, our policy achieves a higher success rate (91.35\%) than any other combination. The constraint helps in reducing the rate of toppled objects to almost half. The policies with the object centroid as EE reach target result in a higher rate of toppled objects (7.73\%) or do not converge at all when combined with the balancing constraint. In the latter case, the arm EE is guided toward the centroid, leading to more violations of the object balance constraint at the start of training, which prevents the agent from discovering the task reward. Therefore, guiding the robot towards interaction with all possible parts of the object surface helps achieve the task.
\begin{table}
\caption{Success rate and percentage of toppled objects in simulation across 4096 simulation runs.}
\label{tab:success_rate_topple}
\vspace{-10pt}
\begin{center}
  \begin{tabular}{p{4.2cm}|P{1.2cm}|P{1.9cm}}
  \hline
   \multirow{2}{4.2cm}{\centering \textbf{Approach}} & \textbf{Success rate [\%]} & \textbf{Toppled object rate [\%]} \\
   \hhline{=|=|=}
   Ours w/o sampling on object surface & 49.80 & 4.50 \\ \hline
  Ours w/o sampling on object surface \& object balance constraint & \multirow{2}{1.2cm}{\centering 88.70} & \multirow{2}{1.9cm}{\centering 7.73} \\ \hline
  Ours w/o object balance constraint                                            & 90.00 & 6.93 \\ \hline 
   Ours (with all the above)                                                           & \textbf{91.35} & \textbf{3.46} \\ \hline

    \end{tabular}
\end{center}
\end{table}

\subsection{Constraint satisfaction}
\label{sec:constraint_satisfaction}
\begin{table}
\caption{Average proportion of time (\%) that each constraint is violated in simulation across 4096 runs}
\label{tab:sim_constaint_violation}
\vspace{-10pt}
\begin{center}
  \begin{tabular}{p{0.5cm}|p{0.5cm}|p{0.5cm}|p{0.5cm}|p{0.5cm}|p{0.5cm}|p{0.6cm}|p{0.5cm}|p{0.5cm}}
  \hline
    $\pmb c_{a}^{base}$ & $\pmb c_{a}^{arm}$ & $\pmb c_{\dot{a}}^{arm}$ & $\pmb c_{q_j}$ & $\pmb c_{\dot{q}_j}$ & $\pmb c_{\tau_j}$ & $\pmb c_{\tau_{j,leg}}$ & $\pmb c_{coll}$ & $c_{\theta_{obj}}$\\
   \hhline{=|=|=|=|=|=|=|=|=}
   0.059 & 0.014 & 0.032 & 0.011 & 0.007 & 0.189 & 0.473 & 0.01 & 0.298 \\ \hline
\end{tabular}
\end{center}
\end{table}

We provide more insights on the achieved constraint satisfaction. We compute the average time proportion for which each constraint is violated during the simulation of an episode with 4096 environments. As shown in \cref{tab:sim_constaint_violation}, all the constraints are violated less than 1\% of the simulated time. The leg joint torque constraint $\pmb c_{\tau_{j,leg}}$ with the higher constraint violation refers to violations of the nominal torque limits and mainly concerns the used locomotion policy. A possible reason for this is that there was no simulated force on the arm EE during training of the locomotion policy.

On the hardware, actuation limits comprise the most challenging constraints to satisfy, in particular, the position limit for the shoulder flexion-extension joint (arm's second joint). The pushing task requires the robot's arm to reach low, especially for thin objects, while the position limit for this joint is slightly larger than 90 deg (with the zero on the positive $z$ axis of the base frame). In practice, this requires the controller to operate this joint very close to the limit during most of the task. Nevertheless, this was achieved during the extensive tests that we carried out.

%
%

\subsection{Robustness against unknown objects \& contact switching}
\label{sec:robustness}

We extensively test the controller on the hardware to move and reorient different objects. The tests were conducted on the protective floor mats of our testing area, which have high friction and can even exhibit small gaps along the mat's seams. Although this increases difficulty, we opted for more challenging conditions that can resemble real-world scenarios. We present the success rates for these challenging tests. Below, we describe the main experiments.

\begin{table}
\caption{Success rate during hardware experiments with objects of different material: plastic (p), cardboard (c), wood (w) and different shape: cuboid (cu), cylinder (cy)}
\label{tab:success_rate_hw}
\vspace{-10pt}
\begin{center}
  \begin{tabular}{P{0.7cm}|P{0.5cm}|P{1.1cm}|p{0.5cm}|P{1.8cm}|P{1.1cm}}
  \hline
   \multirow{2}{0.7cm}{\centering \textbf{Object}} & \textbf{Mass [kg]} & \textbf{Size [cm$^3$]} & \textbf{$\Delta\theta_z$ [deg]} & \textbf{\# of face switches / goal} & \textbf{Success rate [\%]} \\
   \hhline{=|=|=|=|=|=}
   \scriptsize P-CU & 6.43  & 60x34x40           & 180 & 0.90  & 91.6  \\ \hline
   \scriptsize C-CU & 5.30   & 50x50x53           & 0   & 0.23 & 92.9  \\ \hline
   \scriptsize C-CU & 8.32  & 50x50x53           & 90  & 0.75 & 83.3  \\ \hline
   \scriptsize C-CU & 4.5 & 100x50x53          & 0   & 0.14 & 80.0     \\ \hline
   \scriptsize W-CU & 6.30   & 40x40x60           & 180 & 1.00  & 91.6  \\ \hline
   \scriptsize C-CU & 13.30  & 50x50x60           & 0   & 4.80  & 83.3  \\ \hline
   \scriptsize C-CY & 2.45   & \text{$\Phi$}30x40 & 0   & -    & 83.3  \\ \hline
    \end{tabular}
\end{center}
\end{table}

\textit{Success}: This set of experiments consists of sequentially moving and reorienting the object between two fixed goal poses in the space, as shown in \cref{Fig:experiments_ptp}. We do not move the robot manually before sending a new object goal; the policy successfully moves the robot to the appropriate side of the object to push in the correct direction. We tested the learned controller with objects of varying mass, size, shape, and material (\cref{tab:success_rate_hw}), with yaw angle differences $\Delta \theta_z$ of $0^{\circ}$, $90^{\circ}$, or $180^{\circ}$ between the two goal poses. As goals are sent sequentially, the yaw angle difference between the object and the new goal matches $\Delta \theta_z$ (± the success tolerance). The policy achieves a success rate of at least 80\%. For the cuboids, we report in \cref{tab:success_rate_hw} the average contact face switches per goal. A face switch is considered when the robot switches contact to a different face of the cuboid. It can be seen that for higher object yaw orientation changes $\Delta \theta_z$, the controller makes more contact and face switches to properly align the object with the goal, while for $\Delta \theta_z=0$ the robot can most of the time achieve the task without face switch. For cylindrical shapes, a face switch cannot be defined; contact switching is still observed in any case. We demonstrate how the controller manages to move these objects in the accompanying material. In our tests, we also include a 13.3 kg heavy cuboid on caster wheels, although the controller was never trained with wheeled kinematics. In this case, the object's motion overshoots, resulting in more time and contact face switches before ultimately succeeding.

\begin{figure}
  \centering
  \graphicspath{{figures/}}
  \includegraphics[width=\columnwidth]{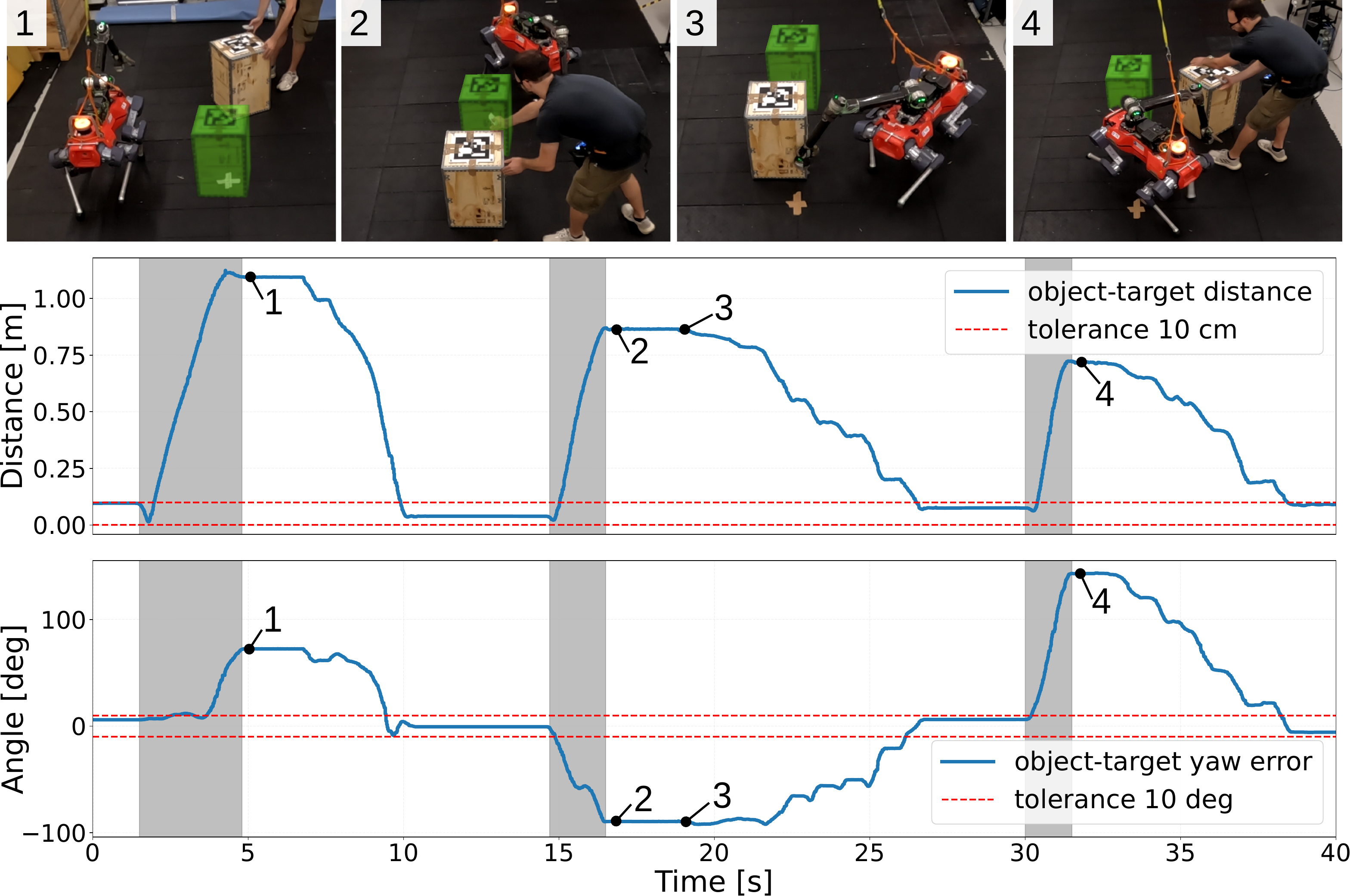}
  \vspace{-22.5pt}
  \caption{Distance and yaw angle error between object and goal. The shaded regions denote the time when the object is moved away from the goal.}
  \label{Fig:experiments_disturbance}
\end{figure}

\textit{Reactive behavior}: In this experiment, we keep the object goal pose fixed in space and move the object away from it multiple times while the policy continuously controls the robot. We repeat the experiment for the objects in \cref{tab:success_rate_hw} and include the motions in the accompanying video. The distance and yaw angle error from one of the objects (W-CU) is shown in \cref{Fig:experiments_disturbance}. The robot successfully pushes the object back to the goal within the specified tolerance. 

In most of the failure cases, the robot first pushes the object very close to the success margin and then stops pushing. We believe that this is not a limitation of the method but rather can be mitigated through further policy tuning.
%

\subsection{Adaptability to object size}
\label{sec:adaptation}
\begin{figure}
  \centering
  \graphicspath{{figures/}}
  \includegraphics[width=0.9\columnwidth]{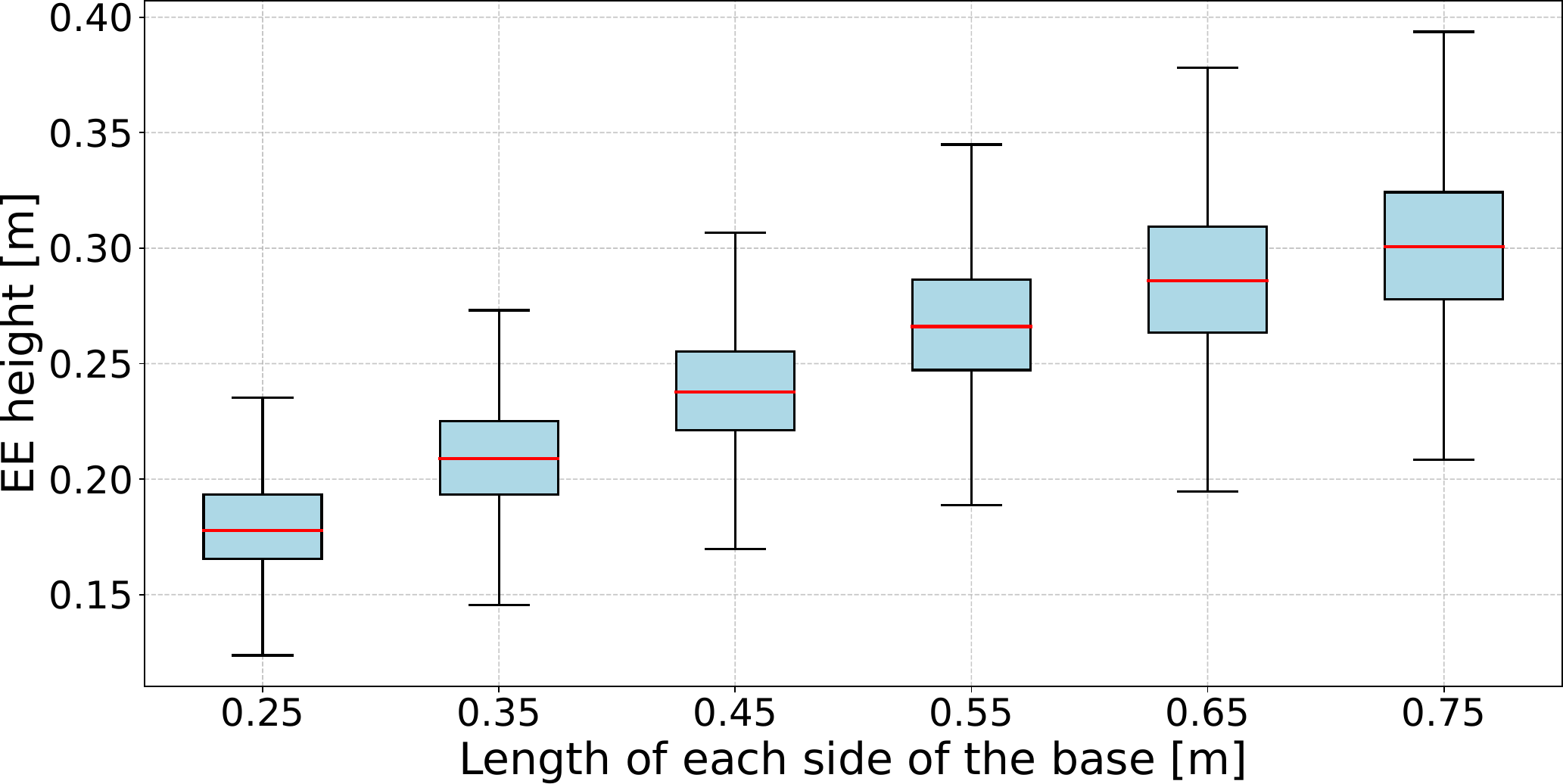}
  \vspace{-8pt}
  \caption{Boxplot of the arm EE height for objects with rectangular bases of different sizes. The policy pushes lower for objects with smaller base.}
  \label{Fig:adaptive_object_xy}
\end{figure}

\begin{figure}
  \centering
  \graphicspath{{figures/}}
  \includegraphics[width=\columnwidth]{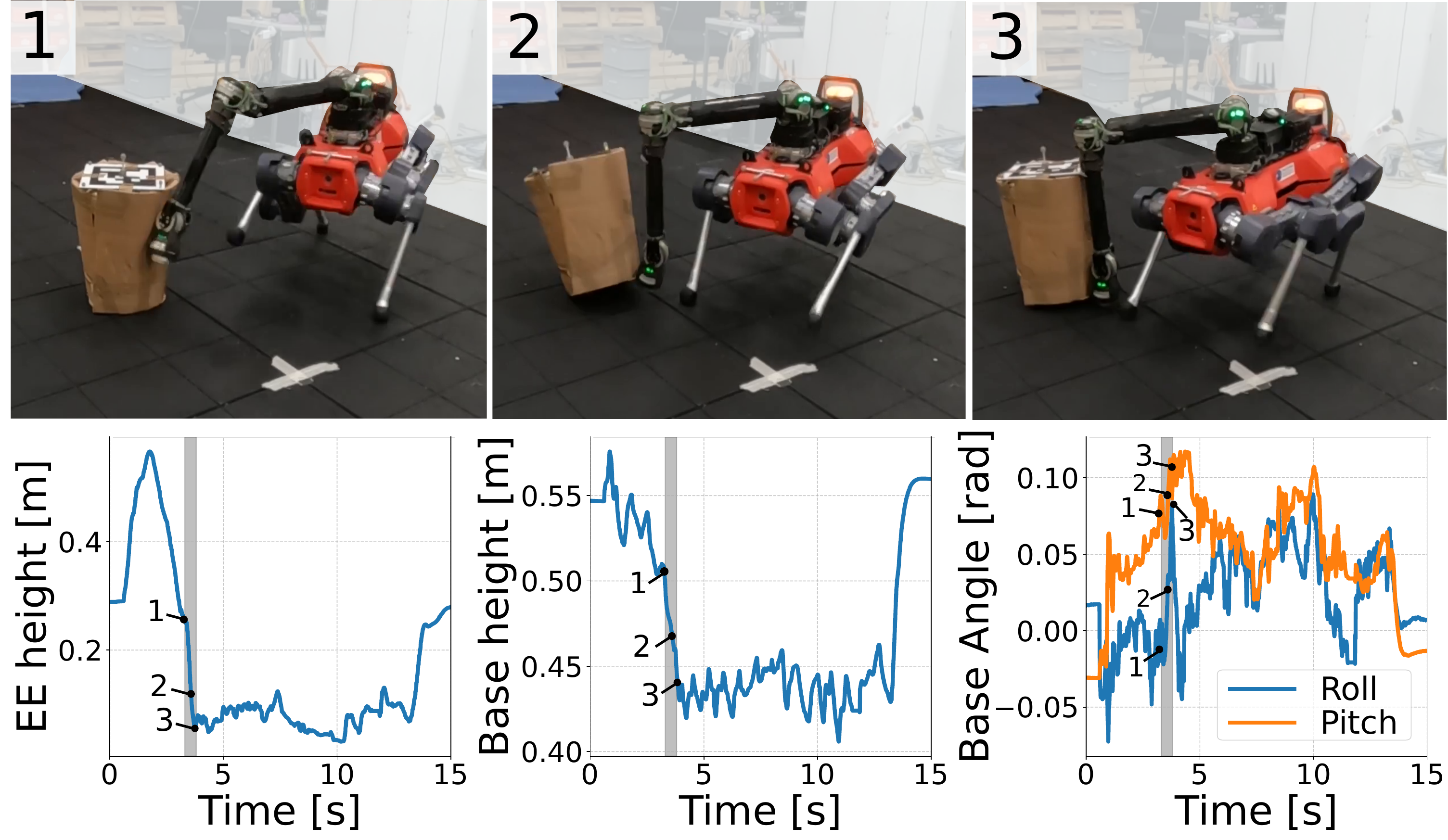}
  \vspace{-20pt}
  \caption{Arm EE height, base height, and orientation while pushing a thin cylinder. The shaded region consists of the time when the cylinder is tilted due to an initial push. The base height and orientation (pitch down, roll) contribute towards pushing power immediately after the object tilts.}
  \label{Fig:cylinder_experiment}
\end{figure}

As mentioned in \cref{sec:domain_randomization}, the dimensions of the objects during training were randomized. We investigate the adaptability of the policy with respect to the object's xy-footprint since thin objects can be prone to toppling. To that end, we select six object xy-footprint sizes equally distributed across the training range and simulate the policy for 1000 successful episodes per size. We fix the object height, mass, center of mass, and friction values to constant and disable the additive observation noise to evaluate the effect of the object's base.  In \cref{Fig:adaptive_object_xy}, we report the height distribution of the robot EE expressed in the world frame while in contact with the object. The policy learns to push lower for objects that have smaller bases. It is of particular interest that the policy does not observe any explicit information regarding the object size or dynamics. This implies that the robot's adaptive behavior is based on the object pose observation. In practice, the robot adapts the pushing location to lower when the object is inclined. We observed such behavior while testing on the real hardware. As shown in \cref{Fig:cylinder_experiment}, the policy can approach and push a thin cylinder on a flooring of high-friction mats. When the cylinder starts tilting, the robot reactively changes the pushing location to lower and avoids object toppling. \cref{Fig:cylinder_experiment} shows that reaching that low with the arm EE is possible due to the contribution of the base motion. In particular, immediately after and during the object tilting (shaded area), the base height, pitch, and roll are adapted to enable reaching lower with the arm EE. This highlights the advantage of using a 6D locomotion policy.
%
%

\section{Conclusion}
We proposed a constrained RL-based controller for dynamically moving and reorienting objects with a mobile manipulator. The controller was extensively tested on hardware and was found to solve the task successfully. The generated motion behaviors are characterized by online contact switching and robustness concerning unknown objects of different mass, size, and shape on a high friction floor. The rate of toppled objects is reduced through an appropriate object balance constraint. By only relying on object pose information, the controller changes the object pushing location to lower for thin objects that may topple. Future directions include adding memory to the policy architecture and using an on-board solution for the perception of the objects.
\section*{ACKNOWLEDGMENTS}
We gratefully acknowledge the advice of Yuntao Ma for policy deployment, Takahiro Miki for the locomotion policy, and Elliot Chane-Sane for the constrained PPO \cite{chanesane2024cat}.



\bibliographystyle{IEEEtran}
\bibliography{bibliography}

\end{document}